\documentclass[letterpaper, 10 pt, conference]{ieeeconf}  

\IEEEoverridecommandlockouts                              
\usepackage{graphicx}
\usepackage{authblk}

\title{\LARGE \bf
Towards a Realistic Simulation Framework for Vehicular Platooning Applications
}

\begin{document}

\author{
   
    Bruno Vieira $^1$, Ricardo Severino $^1$, Anis Koubaa $^{1,2}$, Eduardo Tovar $^1$\\
    $^1$ CISTER Research Centre, ISEP, Polytechnic Institute of Porto\\
    $^2$ Prince Sultan University, Saudi Arabia\\
    \{bffbv, rarss, emt\}@isep.ipp.pt, akoubaa@psu.edu.sa
}

\maketitle
\thispagestyle{empty}
\pagestyle{empty}

\begin{abstract}

Cooperative vehicle platooning applications increasingly demand realistic simulation tools to ease their validation, and to bridge the gap between development and real-word deployment. However, their complexity and cost, often hinders its validation in the real-world. In this paper we propose a realistic simulation framework for vehicular platoons that integrates Gazebo with OMNeT++ over Robot Operating System (ROS) to support the simulation of realistic scenarios of autonomous vehicular platoons and their cooperative control.

\end{abstract}

\section{Introduction}

Vehicular Platooning (VP) is an emerging application of the new generation of safety-critical Cooperating Cyber Physical Systems. Although VP can increase fuel efficiency and road capacity, by having vehicle groups traveling close together, VP presents several safety challenges, considering it heavily relies on wireless communications, and upon a set of sensors that can be affected by noise. The ETSI ITS-G5\cite{c11} is considered as the enabler ready-to-go communications technology for such applications, and although there has been extensive analysis of its performance \cite{c9,c10, c12}, the understanding of its impact upon the safety of these Systems of Systems (SoS) is rather immature. Hence, extensive testing and validation must be carried out to understand the safety limits of such SoS by encompassing communications. However, the expensive equipment and safety risks involved in testing, demands for comprehensive simulation tools that can as accurate as possible mimic the real-life scenarios, from the autonomous driving perspective as well as from the communications perspective. The Robotic Operating System (ROS) framework is already widely used to design robotics applications, and aims at easing the development process by providing multiple libraries, tools and algorithms, and a publish/subscribe transport mechanism. On the other hand, several network simulators are available and capable of carrying out network simulation of vehicular networks. Nonetheless, these remain mostly separated from the autonomous driving reality offering none or very limited capabilities in terms of evaluating cooperative autonomous driving systems. In this work, we carried out the integration of a well-known ROS-based robotics simulator (Gazebo) with a network simulator (OMNeT++), enabling a powerful framework to test and validate cooperative autonomous driving applications. Currently, most relevant simulation frameworks focus on enabling an integration between traffic and network simulators, supporting the evaluation of Inteligent Traffic Systems (ITS), e.g., VSimRTI [1], Artery [3]. Some support vehicular platooning applications, such as Webots[4], VISSIM[6], CORSIM[7]. However, all these come short in comparison with the advanced simulation and capabilities of a robotics simulation framework such as Gazebo, which is developed from scratch to enable a realistic simulation environment for autonomous systems via accurate physical modelling of sensors, actuators and vehicles, while harnessing the power of the ROS development environment. Plexe [2], extension of Veins, aims at enabling platooning simulation, by integrating OMNeT++, SUMO [5] together with a few control and engine models. However, similarly to previous examples, it lacks the power of ROS and only enables the test of longitudinal platooning i.e., no lateral control, and lacks support of a ITS-G5 communication stack, the standard for C-ITS applications in Europe. Hence, this work, is the first to integrate ROS and a network simulator supporting a full stack of the ETSI ITS-G5.

\section{Framework Architecture}

Our simulation framework builds over the Veins simulator and the Vanetza communications’ stack implementation, borrowing and extending much of the middle-ware components from the Artery framework. It relies on ROS publish/subscribe mechanisms to integrate OMNeT++ with Gazebo, represented by blue arrows at Figure 1. Each OMNeT++ node represents a car’s network interface and contain a Vehicle Data Provider (VDP) and a Robot Middleware (RM). VDP is the bridge that supplies RM data from the Gazebo simulator. RM uses this data to fill ITS-G5 CAM’s data fields (i.e Heading, Speed values) through the CaService that proceeds to encode this data fields in order to comply with ITS-G5 ASN-1 definitions. RM also provides GPS positions to position the nodes in the INET mobility module. Regarding synchronization, OMNeT++ is an event-driven simulator and Gazebo is a time-driven simulator, thus synchronizing both simulators represented a challenge. A synchronization module was implemented in OMNeT++, relying on ROS “/Clock” topic as clock reference, which schedules a custom made OMNeT++ message for this purpose (“syncMsg”) to an exact ROS time, forcing the OMNeT++ simulator engine to generate an event upon reaching that timestamp.

\section{Experimental Results}

   \begin{figure}
    \includegraphics[scale=0.35]{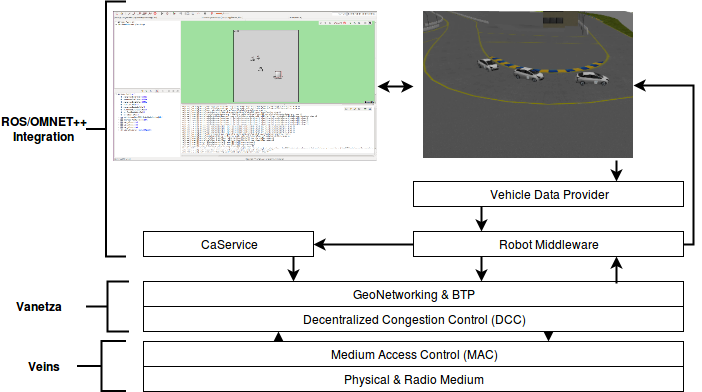}
       \caption{Framework Architecture}
      \label{figurelabel}
   \end{figure}
   
The simulation is composed of three vehicles modeled from a Toyota Prius running a PID-based platooning control model \cite{c10} that solely relies on Cooperative Awareness Messages (CAM) to maintain the platooning service, with a safe distance set to 8 meters. Different CAM sending frequencies were evaluated (10 Hz - 2.5 Hz). At the lower frequency value, the second car fails to keep up with the leader vehicle. Fig. 2 presents a quick overview on how data flows from carX’s sensors into carY’s control application, following a CAM transmission between different nodes in OMNeT++. NodeX and nodeY represent the vehicle’s network interfaces. We analyzed the impact of different CAM exchanging frequencies (Fig.3) on the platoon-following behavior of the second car, regarding the longitudinal distance and steering angles. This provides us with a good perception on how the different frequencies affect the PID longitudinal control of the car. Earlier iteration values confirm that the platooning starts from parked position, and the follower only starts platooning after the leader starts moving, thus the follower needs to accelerate to catch up to its leader. It is also clearly noticeable, in the 0.4s period around iteration 700, the car lost track of the leader vehicle, making a stop and the leader kept going forward. At higher CAM sending frequencies, we can observe that the PID controller shows better stability, and the inter-distance stability improves. The same is visible for the steering behavior. Again, like in the previous graph, around the 700 iterations mark, we can notice the steering angle of the car staying at zero, as it lost track of its leader and stopped. Regarding the other runs on different frequencies, we can get a similar comparison to the previous one, where we can, notice that the PID steering control becomes more stable the higher the CAM sending frequency.

    \begin{figure}
    \centering
    \includegraphics[scale=1.05]{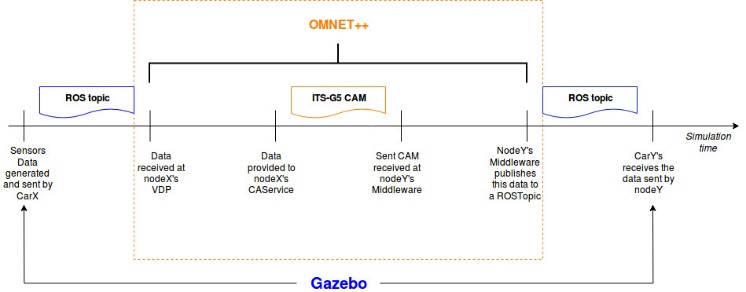}
       \caption{Data workflow}
      \label{figurelabel}
   \end{figure}
   
   \begin{figure}
   \centering
    \includegraphics[scale=0.7]{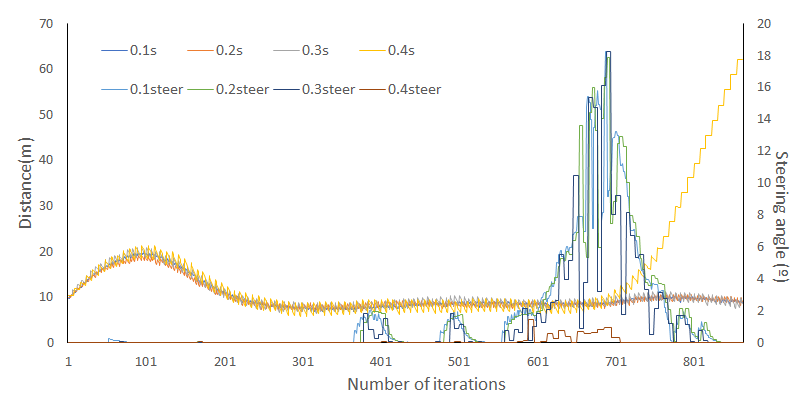}
       \caption{Vehicle inter-distances and steering angles}
      \label{figurelabel}
   \end{figure}

\section{Conclusions and future work}

Initial validation confirms a positive feedback between the network simulation parameters and its effect in the platooning control model. Further validation is to be carried out regarding the impact of the introduced simulator delay, impact of different network parameters and the performance of other platooning control models.
\section{Acknowledgments}  

This work was partially supported by National Funds through FCT/MCTES (Portuguese Foundation for Science and Technology), within the CISTER Research Unit (UID/CEC/04234); also by the FCT and the EU ECSEL JU under the H2020 Framework Programme, within project(s) ECSEL/0002/2015, JU grant nr. 692529-2 (SAFECOP) and ECSEL/0006/2015, JU grant nr. 692455-2 (ENABLE-S3).

\addtolength{\textheight}{-12cm}   







\begin{thebibliography}{99}

\bibitem{c1} B. Schuenemann, “V2X Simulation Runtime Infrastructure VSimRTI: An Assessment Tool to Design Smart Traffic Management Systems” , Computer Networks, Volume 55, pp. 3189-3198, October 2011.
\bibitem{c2} M. Segata,et al., "PLEXE: A Platooning Extension for Veins," Proceedings of 6th IEEE VNC 2014, Dec 2014, pp. 53-60.
\bibitem{c3} R. Riebl, et al. , "Artery: Extending Veins for VANET applications," Int. Conf. Models and Technologies for Intelligent Transportation Systems (MT-ITS), Budapest, 2015.
\bibitem{c4} Michel, O. Webots: Professional Mobile Robot Simulation, Int. Journal of Advanced Robotic Systems, Vol. 1, Num. 1, pages 39-42.
\bibitem{c5} P. A. Lopez et al., "Microscopic Traffic Simulation using SUMO," 2018 21st International Conference on Intelligent Transportation Systems (ITSC), Maui, HI, 2018, pp. 2575-2582.
\bibitem{c6} VISSIM: microscopic, behavior-based multi-purpose traffic simulation program. http://www.ptvamerica.com/vissim.html.
\bibitem{c7} L. Owen et al., “Traffic Flow Simulation Using CORSIM,” Proc.2000 Winter Simulation Conf., 2000, pp. 1143–147.

\bibitem{c9} O. Karoui, et al. , Mohamed Khalgui, Anis Koubâa, Emna Guerfala, Zhiwu Li, Eduardo Tovar, “Dual mode for vehicular platoon safety: Simulation and formal verification”, Information Sciences, Volume 402, 2017, Pages 216-232
\bibitem{c10} O. Karoui, et al. , “Performance evaluation of vehicular platoons using Webots”, IET Intelligent Transport Systems, Volume 11, Issue 8, 201
\bibitem{c11}ETSI, "ETSI EN 302 665 V1.1.1(09-2010):" Intelligent Transport Systems (ITS);
Communications Architecture"
\bibitem{c12} A. Koubaa et al. , 'System and method for operating a follower vehicle in a vehicle platoon ', US9927816B2, 2016



\end{thebibliography}
\end{document}